\documentclass[sigconf]{acmart}

\AtBeginDocument{%
  }

\copyrightyear{2025}
\acmYear{2025}
\setcopyright{acmlicensed}
\acmConference[MM '25]{Proceedings of the 33rd ACM International Conference on Multimedia}{October 27--31, 2025}{Dublin, Ireland}
\acmBooktitle{Proceedings of the 33rd ACM International Conference on Multimedia (MM '25), October 27--31, 2025, Dublin, Ireland}
\acmISBN{979-8-4007-2035-2/2025/10}
\acmDOI{10.1145/3746027.3754772}

\definecolor{cvprblue}{rgb}{0.21,0.49,0.74}
\usepackage{balance}
\usepackage{epsfig}
\usepackage{graphicx}
\usepackage{amsmath}

\usepackage{booktabs}
\usepackage{multirow}
\usepackage{tabularx}
\usepackage{enumitem}
\usepackage{gensymb}
\usepackage{colortbl}
\usepackage{caption}
\usepackage{makecell}
\usepackage{wrapfig}

\usepackage{pifont}
\definecolor{mycyan}{cmyk}{.1,0,0,0}
\definecolor{mygray}{gray}{.95}
\newcommand{\mypara}[1]{\vspace{1mm}\noindent\textbf{#1}}
\usepackage[capitalize]{cleveref}

\newcommand{\name}{OmniGen}
\newcommand{\UAE}{UAE}

\newcommand{\tr}[2][\@empty]{%
  \ifx#1\@empty
    \textcolor{red}{#2}%
  \else
    \textcolor{red}{\fontsize{#1}{#1}\selectfont #2}%
  \fi
}
\newcommand{\tb}[2][\@empty]{%
  \ifx#1\@empty
    \textcolor{blue}{#2}%
  \else
    \textcolor{blue}{\fontsize{#1}{#1}\selectfont #2}%
  \fi
}
\newcommand{\ts}[2][\@empty]{%
  \ifx#1\@empty
    {#2}%
  \else
    {\fontsize{#1}{#1}\selectfont #2}%
  \fi
}

\usepackage{colortbl}

\usepackage{bbding}
\definecolor{minetable1colorx}{rgb}{0.75, 0.75, 0.75}

\definecolor{mycyan}{cmyk}{.1,0,0,0}
\definecolor{mygray}{gray}{.95}
\definecolor{mypink}{rgb}{.99,.91,.95}

\newcommand{\TODO}[1]{\textcolor{red}{\textbf{[TODO: not finish or just a placeholder]}}}

\newcommand{\mb}[1]{\mathbf{#1}}

\usepackage{bbm}

\begin{document}

\title{\name{}: Unified Multimodal Sensor Generation for Autonomous Driving}

\author{Tang Tao}
\authornote{Work done during an internship at Li Auto Inc.}
\orcid{0000-0001-8526-220X}
\affiliation{
  \institution{Shenzhen Campus of Sun Yat-sen University}
  \city{Shenzhen}
  \country{China}
}
\email{trent.tangtao@gmail.com}

\author{Enhui Ma}
\affiliation{%
	\institution{Westlake University}
  \city{Hangzhou}
  \country{China}
}
\email{
maenhui@westlake.edu.cn
}

\author{Xia Zhou}
\affiliation{%
  \institution{Li Auto Inc.}
  \city{Beijing}
  \country{China}
}
\email{zhouxia@lixiang.com}

\author{Letian Wang}
\affiliation{%
	\institution{The University of Toronto}
  \city{Toronto}
  \country{Canada}
}
\email{
letianwang0@gmail.com
}

\author{Tianyi Yan}
\affiliation{%
  \institution{University of Macau}
  \city{Macau}
  \country{China}
}
\email{tianyi.yan123@gmail.com}

\author{Xueyang Zhang}
\affiliation{%
  \institution{Li Auto Inc.}
  \city{Beijing}
  \country{China}
}
\email{zhangxueyang@lixiang.com}

\author{Kun Zhan}
\affiliation{%
  \institution{Li Auto Inc.}
  \city{Beijing}
  \country{China}
}
\email{zhankun@lixiang.com}

\author{Peng Jia}
\affiliation{%
  \institution{Li Auto Inc.}
  \city{Beijing}
  \country{China}
}
\email{jiapeng@lixiang.com}

\author{Xianpeng Lang}
\affiliation{%
  \institution{Li Auto Inc.}
  \city{Beijing}
  \country{China}
}
\email{langxianpeng@lixiang.com}

\author{Jia-Wang Bian}
\affiliation{%
  \institution{Bytedance Seed}
  \city{San Jose}
  \country{United States}
}
\email{jiawang.bian@gmail.com}

\author{Kaicheng Yu}
\affiliation{%
  \institution{Westlake University}
  \city{Hangzhou}
  \country{China}
}
\email{kaicheng.yu.yt@gmail.com}

\author{Xiaodan Liang}
\authornote{Corresponding author.}
\affiliation{%
	\institution{Shenzhen Campus of Sun Yat-sen University}
  \city{Shenzhen}
  \country{China}
}
\email{hxdliang328@gmail.com}



\renewcommand{\shortauthors}{Tao Tang, et al.}

\begin{abstract}
Autonomous driving has seen remarkable advancements, largely driven by extensive real-world data collection. However, acquiring diverse and corner-case data remains costly and inefficient. Generative models have emerged as a promising solution by synthesizing realistic sensor data. However, existing approaches primarily focus on single-modality generation, leading to inefficiencies and misalignment in multimodal sensor data. 
To address these challenges, we propose \name{}, which generates aligned multimodal sensor data in a unified framework. 
Our approach leverages a shared Bird’s Eye View (BEV) space to unify multimodal features and designs a novel generalizable multimodal reconstruction method, \UAE{}, to jointly decode LiDAR and multi-view camera data.
\UAE{} achieves multimodal sensor decoding through volume rendering, enabling accurate and flexible reconstruction.
Furthermore, we incorporate a Diffusion Transformer (DiT) with a ControlNet branch to enable controllable multimodal sensor generation.
Our comprehensive experiments demonstrate that \name{} achieves desired performances in unified multimodal sensor data generation with multimodal consistency and flexible sensor adjustments.
\end{abstract}



\begin{teaserfigure}
  \includegraphics[width=\textwidth]{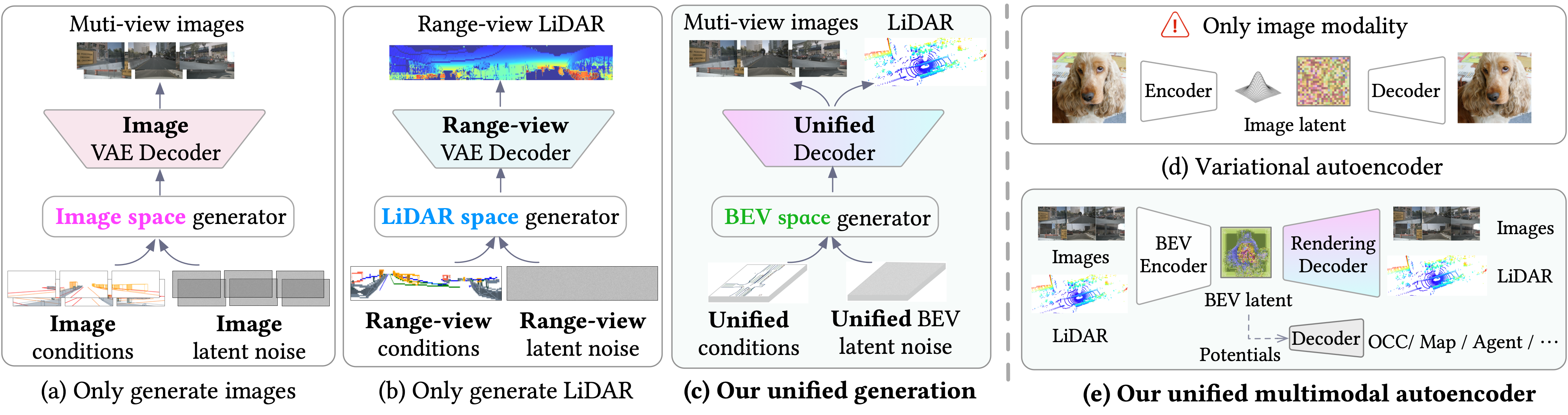}
  \caption{(a)(b) Existing approaches primarily
focus on single-modality generation on its own space, leading to inefficiencies and
misalignment in multimodal sensor data. (c) Our \name{}, a unified multimodal sensor generation framework. (d) Traditional VAE, which only supports images.  (e) Our \UAE{}, a unified multimodal autoencoder, leverages the unified BEV latent.} 
  \label{fig:teaser}
\end{teaserfigure}



\maketitle

\section{Introduction}
\label{sec:intro}
Autonomous driving has made remarkable progress in recent years, driven by large-scale real-world data collected from diverse environments. However, the high cost and inefficiency of acquiring real-world data, especially in corner-case scenarios, limit the scalability of this process.
Meanwhile, generative models~\cite{zhang2023adding, peebles2023scalable, brooks2024video} have gained significant attention for their ability to learn data distributions and synthesize realistic content, achieving remarkable success in image generation.
Consequently, using generative models to synthesize desired sensors has become a de-facto standard in autonomous driving to address the data scarcity issue.
For camera data generation, researchers directly fine-tune existing image generation models, such as ControlNet~\cite{zhang2023adding}, with specific driving scene layouts or textual descriptions as conditions to generate scene images.
Moreover, beyond cameras, LiDAR sensors play a crucial role in practical autonomous driving systems for accurate perception and planning, providing reliable 3D environmental measurements by capturing point clouds.
For LiDAR data generation, to leverage existing image generation models, researchers first convert point clouds into range-view pseudo images and then adapt the whole image generation pipeline, \textit{i.e.}, the autoencoder and diffusion model, to range-view space to generate pseudo LiDAR images, which are subsequently converted back into LiDAR points.

However, as illustrated in ~\cref{fig:teaser}, existing driving scene generation models primarily focus on single-modality sensor data, while the unified generation of multimodal data remains unexplored.
Unified multimodal sensor generation offers several advantages:
\textbf{Improved efficiency} – generating both modalities simultaneously eliminates the need for separate pipelines (\textit{e.g.}, the data processing, model training, and model updates).
\textbf{Better sensor alignment} – Independently generating sensor data is difficult to align across different sensors, making it challenging for downstream multimodal models to utilize effectively.
Nevertheless, achieving a unified multimodal generation poses significant challenges. Camera generation models operate in the image latent space, where conditions are projected into the perspective image view, while LiDAR generation models generate data in the range-view latent space with conditions projected accordingly. 
Fusing these distinct generation spaces into a unified representation, while ensuring it is controllable under a unified condition, is a non-trivial problem.

In this paper, we introduce \name{}, a unified multimodal sensor generation framework for autonomous driving. We address the challenges of unified sensor generation by breaking the problem down into several key steps:
\textbf{1) Establishing a unified representation space}: Inspired by prior multimodal perception research~\cite{liu2023bevfusion, li2022bevformer}, such as BEVFusion~\cite{liang2022bevfusion}, we unify multimodal features in a shared Bird's Eye View (BEV) space, which 
provides a global scene context and aligns well with conditions such as textual descriptions or road sketches.
\textbf{2) Decoding multimodal sensor data from the unified BEV space}: Drawing inspiration from generalizable NeRF approaches, \textit{e.g.,} PixelNeRF~\cite{yu2021pixelnerf}, we leverage volume rendering to render sensor data. 
While previous generalizable NeRF methods primarily focus on object-level reconstruction and single-modality image rendering, their application to large-scale and multimodal autonomous driving scenes remains limited. 
Thanks to the unified BEV representation, in this work, we introduce \UAE{}, the first generalizable LiDAR-camera multimodal reconstruction method for autonomous driving scenes. 
To decode multimodal data, \UAE{} first upsamples the BEV features into 3D voxel features and then renders sensor features by sampling the voxel features. 
Subsequently, a carefully designed feature decoder is incorporated to map the rendered features to the corresponding sensor data, preserving and enhancing high-frequency details for improving reconstruction quality.
Compared to traditional VAEs~\cite{kingma2013auto}, \UAE{} naturally offers multiple benefits from its generalizable reconstruction capability.
For instance, autonomous driving scenes typically involve multiple surrounding images. Previous works process each view independently using VAEs and apply attention modules as soft constraints to enforce cross-view consistency. 
In contrast, \UAE{} inherently supports multi-view consistent rendering by leveraging unified BEV features as a global scene constraint.
Similarly, multimodal sensor data rendered from the shared BEV features also maintains consistency across different modalities.
Moreover, \UAE{} allows for the adjustment of sensor parameters (both intrinsic and extrinsic) during the decoding process. This enables effortless camera control, which previously required complex designs and training of generation models to achieve~\cite{yao2024mygo, liu2024mvpbev}.
\textbf{3) Generating latent BEV features for multimodal sensor data}: To enable generative models to produce multimodal sensor data, we enhance \UAE{} with a Vector Quantization (VQ) module, and leverage it as our multimodal autoencoder for unified generation.
For the generative model, we employ the ControlNet-Transformer architecture, which incorporates ControlNet into the powerful Diffusion Transformer (DiT) model. Moreover, we employ comprehensive scene conditions, i.e, scene textual descriptions, BEV road sketches, and 3D boxes, enabling more fine-grained and precise control of the generative model to generate the desired latent BEV features effectively.
Overall, we achieve an end-to-end unified multimodal sensor generation framework. Given driving scene conditions, our \name{} can generate aligned multimodal sensor data.
Moreover, we explore various multimodal architecture designs and provide valuable investigations into unified multimodal generation.

Through comprehensive experiments, we validate the effectiveness of our approach in generating multimodal sensor data within a unified framework.
Specifically, our \UAE{} module achieves state-of-the-art performance over previous generalizable reconstruction, and our \name{} achieves comparable results with previous specialized single-modality methods and effectively generates multimodal data with cross-modality alignment and flexible sensor control, further enhancing downstream autonomous driving tasks.

Overall, our contributions are summarized as follows:
\begin{itemize}
    \item We introduce \name{}, a unified multimodal sensor generation framework that enables the controllable generation of aligned LiDAR and multi-view camera data.
    \item We propose \UAE{}, a generalizable multimodal reconstruction method, which serves as an efficient multimodal autoencoder for encoding and decoding multimodal data in a unified space while allowing multimodal and multi-view consistency and flexible sensor adjustments.
    \item We demonstrate the effectiveness of our method quantitatively and qualitatively through extensive experiments conducted on multiple scenes.
\end{itemize}











\section{Related Work}
\label{sec:related}

\subsection{Camera Generation Models}

Recent advancements in generative models, particularly in diffusion models~\cite{ho2020denoising, zhang2023adding}, have inspired the generation of high-fidelity and controllable driving scenes.
Previous works~\cite{jia2023adriver, gao2024vista, lu2023wovogen, lu2024seeing, chen2024unimlvg, wen2023panacea, ma2024unleashing, wang2023driveWM, xie2025glad, gao2024magicdrivedit} have fine-tuned image generation models on driving data, incorporating various control signals such as maps, object bounding boxes, and textual descriptions to generate diverse driving scenarios.
Specifically, BEVGen~\cite{swerdlow2023_bevgen} first introduces the use of a BEV map as a condition for generating multi-view street images.
BEVControl~\cite{yang2023bevcontrol} further proposes a two-stage generation pipeline that integrates cross-view attention.
MagicDrive~\cite{gao2023magicdrive} highlights 3D geometric information, encoding boxes and road maps separately to enable more fine-grained control.
DriveDreamer~\cite{wang2023drivedreamer} and DriveDreamer-2~\cite{zhao2024drivedreamer-2} generate multi-view video data based on diverse control signals.
Recently, researchers have been further refining driving image generation. For example, some works~\cite{yan2024drivingsphere, li2024syntheocc, li2024uniscene}, \textit{e.g.}, InfiniCube~\cite{lu2024infinicube}, adopt semantic occupancy as an intermediate representation to improve generation quality.
Others~\cite{lu2024infinicube, gao2024magicdrive3d, zou2025mudg} incorporate additional reconstruction modules to synthesize 4D driving scenes.
Moreover, some studies~\cite{yao2024mygo, liu2024mvpbev} integrate camera pose parameters into the generator to achieve more precise control.
In this work, beyond scene images, we aim to jointly generate multimodal sensor data within a unified framework.

\subsection{LiDAR Generation Models}
Generative models also provide a promising alternative for creating realistic LiDAR point clouds without physics-based platforms.
Early approaches, such as LiDARVAE and LiDARGAN~\cite{caccia2019deep}, employ VAE or GANs for LiDAR cloud generation, but the realism achieved in their results is relatively limited.
Following, UltraLiDAR~\cite{xiong2023learning} utilizes VQ-VAE~\cite{van2017neural} to generate voxelized LiDAR point clouds, while LidarDM~\cite{zyrianov2024lidardm} employs a map-conditioned diffusion model to generate scene meshes, which are then raycast to produce LiDAR scans.
Meanwhile, with the advanced diffusion models, many works explore LiDAR generation based on range-view image representations.
LiDARGen~\cite{zyrianov2022learning} firstly applies a diffusion model on range-view images, leveraging progress in image diffusion models.
R2DM~\cite{nakashima2023lidar} designs a more mature diffusion framework, achieving significant performance improvements.
RangeLDM~\cite{hu2024rangeldm} further optimizes efficiency and quality by compressing range-view data into a latent space before diffusion.
More recently, LiDM~\cite{ran2024towards} and Text2LiDAR~\cite{wu2024text2lidar} explore conditional LiDAR generation using conditions such as text, bounding boxes, and maps.
Despite these advancements, state-of-the-art LiDAR generation methods operate in range-view space, which is challenging to unify with the image space used for camera sensor generation.
In this work, we address this issue by unifying multimodal features within a shared BEV space, enabling unified multimodal sensor generation.

\subsection{Generalizable NeRFs}
Latent diffusion models fundamentally operate in the latent space of autoencoders.
However, commonly used autoencoders, such as VAE~\cite{kingma2013auto} and VQ-VAE~\cite{van2017neural}, do not inherently support multimodal sensor data.
On the other hand, generalizable NeRFs~\cite{yu2021pixelnerf, chen2021mvsnerf, lin2022efficient, zhang2022nerfusion} replace the costly per-scene optimization with a single feedforward pass.
These models take several images as input and generate corresponding image outputs, effectively functioning as a more versatile autoencoder.
While previous generalizable NeRF methods have primarily focused on object-level reconstruction, only DistillNeRF~\cite{wang2024distillnerf} recently explored generalizable scene reconstruction for driving scenes. However, it relies on multiple complex modules, such as distillation from offline NeRFs, distillation from foundation models, hierarchical octree representation, and integration of depth features from DepthAnything~\cite{yang2024depth}, and it remains limited to image rendering.
A unified and end-to-end multimodal generalizable NeRF—or a multimodal autoencoder—remains unexplored.
In this work, we introduce \UAE{}, a generalizable multimodal reconstruction method for driving scenes. Furthermore, we enhance \UAE{} with a Vector Quantization (VQ) module and leverage it as a multimodal autoencoder for unified sensor data generation.


\begin{figure*}[ht]
    \centering
    \includegraphics[width=\linewidth]{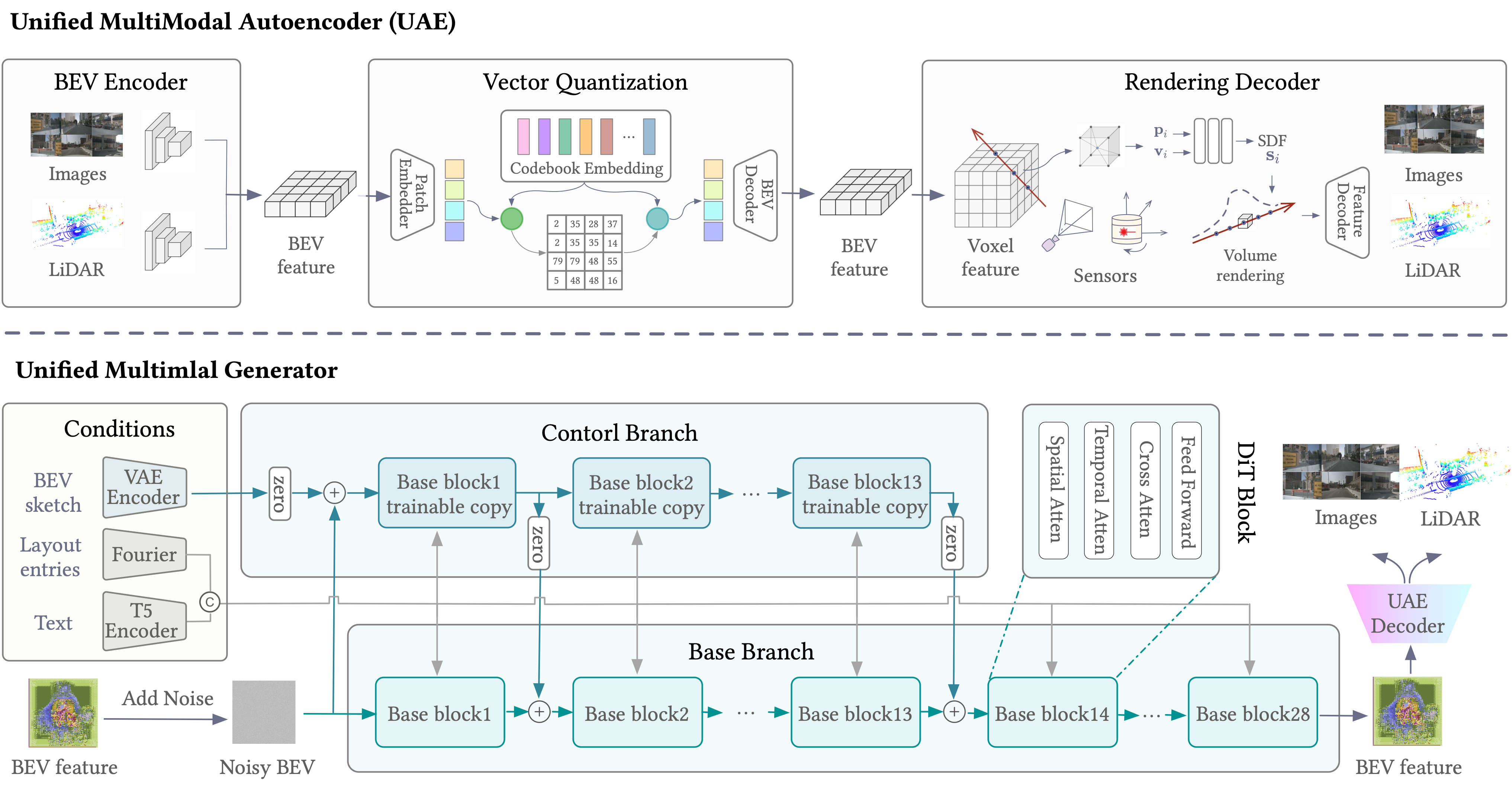}
    \caption{\textbf{Our \name{} framework consists of two main components: the unified multimodal autoencoder and the unified multimodal generator.} The \UAE{} is composed of three parts: a multimodal BEV encoder, a Vector Quantization module, and a multimodal rendering decoder. The unified multimodal generator includes two branches: the base branch and the control branch, which take multiple scene conditions as input to generate BEV latent representations.}
    \label{fig:framework}
\end{figure*}
\section{\name}
\label{sec:method}

In this section, we introduce \name{} in detail. We first give an overview in \cref{subsec: overview}. Then, we introduce the \UAE{} in \cref{subsec: uae} and unified LiDAR-Camera generation in \cref{subsec: unigen}.

\subsection{Overview}
\label{subsec: overview}
As illustrated in \cref{fig:framework}, our \name{} consists of a multimodal autoencoder and a multimodal generator. 
The multimodal autoencoder, \UAE{}, first encodes the camera and LiDAR sensor data into a unified BEV space, then decodes the unified BEV features to multimodal sensor data by volume rendering. 
The multimodal generator adopts ControlNet-Transformer architecture, which incorporates ControlNet into the Diffusion Transformer (DiT) model, enabling fine-grained and precise control.
Given textual descriptions, BEV road sketches, and 3D boxes as scene conditions input, the generator generates the desired latent BEV features effectively.

\subsection{Unified Multimodal Autoencoder} 
\label{subsec: uae}
As commonly used autoencoders, \textit{e.g.}, VAE~\cite{kingma2013auto} and VQ-VAE~\cite{van2017neural}, do not support multimodal sensor data, we propose a novel unified multimodal autoencoder based on volume rendering and dub it as \UAE{}.
The \UAE{} contains three parts: a multimodal BEV encoder, a multimodal rendering decoder, and a Vector Quantization module.

\mypara{Multimodal BEV Encoder.}
The multimodal BEV encoder converts different modalities into a unified BEV space while preserving as much sensor-specific information as possible.
Specifically, following prior works~\cite{liang2022bevfusion, li2022uvtr}, for multi-view images from the camera sensors, we adopt the Lift-Splat-Shoot (LSS) view transformation~\cite{philion2020lift} to lift 2D features into the 3D volume features, denoted as $\mathbf{V}_C\in{\mathbb{R}^{X\times Y\times Z\times C}}$.
For the LiDAR sensor, the point encoder firstly learns a parameterized voxelization~\cite{zhou2018voxelnet} of the raw point clouds and then utilizes sparse 3D convolution networks~\cite{yan2018second} for efficient feature extraction. We follow UVTR~\cite{li2022uvtr} to directly retain the height dimension in the point encoder to obtain LiDAR 3D volume features, $\mathbf{V}_L\in{\mathbb{R}^{X\times Y\times Z\times C}}$.
Then, $\mathbf{V}_C$ and $\mathbf{V}_L$ are summed and passed through a projection layer to enhance the fused voxel representation, forming the unified voxel space $\mathbf{V}_U\in{\mathbb{R}^{X\times Y\times Z\times C}}$.
Finally, we adopt the Spatial-to-Channel (S2C) operation~\cite{xie2022m} to reshape $\mathbf{V}_U$ into the unified BEV space, $\mathbf{B}_U\in{\mathbb{R}^{X\times Y\times (Z\times C)}}$, effectively preserving semantic information while reducing computational cost.
To recover the 3D volume features, we apply the inverse Channel-to-Spatial (C2S) reshaping to the BEV features.
The unified BEV features serve as the target for the generation model, while the voxel features act as the input for the rendering decoder.


\mypara{Multimodal Rendering Decoder.}
When given the unified volumetric features, the multimodal rendering decoder aims to decode multimodal sensor data, which can be divided into two distinct components: image reconstruction and LiDAR reconstruction.
Practically, we represent a scene as an implicit signed distance function (SDF) field from Neus\cite{wang2021neus} as UniPAD~\cite{yang2024unipad}
and use differentiable volume rendering to render multimodal data.

1) For image rendering, we sample camera rays from multi-view images $\mathbf{r}(t) = \mathbf{o} + t\mathbf{d}$ with the camera center $\mathbf{o}$ and viewing direction $\mathbf{d}$.
For each ray $\mathbf{r}_i$, we sample $N$ points $\{\mathbf{p}_{i} = (x_i, y_i, z_i)\}_{i=1}^{N}$ along the ray.
For each sampled point $\mathbf{p}_{i}$, the corresponding features $\mathbf{v}_i$ are obtained from the voxel features $\mathbf{V}_U$ according to its position by trilinear interpolation.
Then, the SDF value $s_i$ is predicted by $\phi_\mathrm{SDF}(\mathbf{p}_i, \mathbf{v}_i)$, where $\phi_\mathrm{SDF}$ represents a shallow MLP.
Then, we render the camera feature descriptor by integrating the sampled features along rays:
\begin{equation}
\mathbf{F}_c = \sum_{i=1}^{N} w_i \mathbf{v_i}, w_i = \alpha_i \prod_{j=1}^{i-1}(1-\alpha_j), \\	\alpha_i=\max \left(\frac{\sigma_s\left(s_i\right)-\sigma_s\left(s_{i+1}\right)}{\sigma_s\left(s_i\right)}, 0\right),
\label{eq:camera}
\end{equation}
where $\sigma_s(x)=(1+e^{-sx})^{-1}$ is a sigmoid function modulated by a learnable parameter $s$.
After obtaining the 2D feature map $\mathbf{F}_c \in \mathbb{R}^{H_f^c \times W_f^c \times C_f}$ for each camera, we design a feature decoder to map the rendered features to the RGB image $\mathbf{I}_c \in \mathbb{R}^{H^c \times W^c \times 3}$ with enhanced high-frequency details.
We employ MSE and LPIPS loss for rendered image supervision, \textit{i.e.}, $\mathcal{L}_{\text{Camera}} = \mathcal{L}_{\text{mse}} + \mathcal{L}_{\text{lpips}}$.

2) For LiDAR rendering, following LiDAR-NeRF~\cite{tao2023lidar}, we treat the oriented LiDAR laser beams as a set of camera rays. Slightly abusing the notation, let $\mathbf{r}(t) = \mathbf{o} + t\mathbf{d}$ be a ray casted from the LiDAR sensor, where $\mathbf{o}$ denotes the LiDAR center, and $\mathbf{d}$ represents the normalized direction vector of the corresponding beam.
Then, similar to the camera rendering, we sample the ray points and get the corresponding features, and render the LiDAR depth measurement and feature descriptor as:
\begin{equation}
D = \sum_{i=1}^{N} w_i t_i,  \mathbf{F}_l = \sum_{i=1}^{N} w_i \mathbf{v_i}.
\label{eq:lidar}
\end{equation}
The rendered depth $D$ can convert into LiDAR points as $ (x, y, z) = (   D \cos(\alpha)\cos(\beta),
  D  \cos(\alpha)\sin(\beta),
   D \sin(\alpha)) $, where $\alpha$ is the vertical rotation and $\beta$ is the horizontal rotation of  viewing direction $\mathbf{d}$.
The rendered LiDAR feature map is also decoded through a feature decoder to the view-dependent features of LiDAR, $\mathbf{I}_l \in \mathbb{R}^{H^l \times W^l \times 2}$, including the intensities and ray-drop probabilities.
We employ L1 loss for depth optimization and MSE loss for intensities and ray-drop supervision, $\mathcal{L}_{\text{LiDAR}} = \mathcal{L}_{\text{depth}} + \mathcal{L}_{i} + \mathcal{L}_{r}$.

Compared to traditional VAEs~\cite{kingma2013auto}, the rendering decoder naturally offers multiple benefits from its generalizable reconstruction capability.
For instance, autonomous driving scenes typically involve multiple surround-view images. Previous works process each view independently using VAEs and apply attention modules as soft constraints to encourage cross-view consistency. However, this approach failed to maintain strict geometric consistency across different views.
In contrast, the rendering decoder inherently supports multi-view consistent rendering by leveraging unified BEV features as a global scene constraint.
Similarly, multimodal sensor data rendered from the shared BEV features also maintains consistency across different modalities.
Furthermore, the rendering decoder allows for flexible adjustment of sensor parameters, \textit{i.e.}, both intrinsic and extrinsic, during the rendering process. This enables intuitive camera control, which previously required complex designs and specialized training to achieve~\cite{yao2024mygo, liu2024mvpbev}.


\mypara{Vector Quantization.}
To better leverage the diffusion model for generating BEV features, we further project the unified BEV features into a tokenized discrete space as VQ-VAE~\cite{van2017neural}, which consists of three modules: BEV Patch Embedder, Vector Quantization, and BEV Feature Decoder. 

1) BEV Patch Embedder. 
We firstly patchify the BEV features $\mathbf{B}_U\in{\mathbb{R}^{X\times Y\times C}}$ into a sequence of BEV patches $\left\{\mathbf{P}^i\in \mathbb{R}^{ P\times P \times C}\right\}_{i=1}^M$, where ${P}$ is the patch size, and $M=H^bW^b/P^2$ is the patch number. Then each BEV patch is further embedded to $\mathbf{z}_c \in \mathbb{R}^{E}$, where $E$ is the embedded dimension.
2) Vector Quantization (VQ). 
We then define a discrete latent space $\{\mathbf{v}_1, \ldots, \mathbf{v}_k, \ldots, \mathbf{v}_K\} \in \mathbb{R}^{K\times E} $ as our codebook embedding, where $K$ represents the maximum number of the embeddings. 
Taking the continuous latent vector $\mathbf{z}_c$ from the patch embedder, the VQ module outputs discrete latent vector $\mathbf{z}_d$ through the nearest neighbor search in the codebook.
3) BEV Feature Decoder. 
We finally feed the discrete BEV embeddings $\{\mathbf{z}_d^{i}\}_{i=1}^{M}$ to our BEV feature decoder by reshaping them into a grid format and then reconstructing the original BEV features. Detailed architecture is present in \cref{fig:UAEdecoder} of the supplementary material.

The overall VQ training loss $\mathcal{L}_{\text{vq}}$ includes the codebook loss $\mathcal{L}_{\text{code}}$ and the reconstruction loss $\mathcal{L}_{\text{re}}$. Due to the non-differentiable vector quantization operation, the codebook loss is defined as:
\begin{equation}
\mathcal{L}_{\text{code}} = \frac{1}{M} \sum_{i=1}^{M} \left(
\begin{Vmatrix}{\mathbf{z}_d^{i} - \verb'sg'(\ell_2(\mathbf{z}_c^{i}))}\end{Vmatrix}_2^{2} + 
\begin{Vmatrix}{\verb'sg'(\mathbf{z}_d^{i}) - \ell_2(\mathbf{z}_c^{i})}\end{Vmatrix}_2^{2}
\right),
\label{eq:code-loss}
\end{equation}
where $\ell_2$ means L2 normalization and $\verb'sg'$ denotes stop-gradient. 
We utilize MSE for the reconstruction loss of BEV features, and the final loss is defined as:
\begin{equation}
\mathcal{L}_{\text{vq}} = \mathcal{L}_{\text{re}} + \mathcal{L}_{\text{code}}.
\label{eq:vq-loss}
\end{equation}

\mypara{Overall Optimization.}
The overall optimization target of our \UAE{} is formulated as:
\begin{equation}
\mathcal{L}_{\text{\UAE{}}} = \mathcal{L}_{\text{Camera}} + \mathcal{L}_{\text{LiDAR}}  + \mathcal{L}_{\text{vq}}.
\label{eq:uae}
\end{equation}

\subsection{Unified LiDAR-Camera Generation}
\label{subsec: unigen}

\mypara{Latent BEV Feature DiT.}
Latent Diffusion Models (LDMs)~\cite{rombach2022high} perform diffusion process on the latent space, using a pre-trained VAE to map between the input and the latent space.
It iteratively denoises from a random Gaussian noise $\mathbf{z}^{c}_R$ for $R$ steps with a denoiser $\mathcal{G}^{c}$ into a clean image latent $\mathbf{z}^{c}_0$.
This VAE+diffusion formulation is widely adopted in image generation, and we also adopt this manner for our unified BEV features generation.
As our BEV features already leverage vector quantization into discrete space, we do not need to apply additional VAE mapping. 
Given scene conditions $\mb{S}$, the goal is to generate corresponding BEV features from latent variables $\epsilon\sim{\mathcal{N}(0, I)}$, \textit{i.e.}, \(\mathbf{B}_{U} = \mathcal{G}(\mathbf{S}, \epsilon\}\).
Then the generated BEV features are decoded to multimodal sensor data through the multimodal rendering decoder of our  \UAE{}.
To enhance the quality of generated BEV features, we adapt the advanced DiT~\cite{opensora} as our denoiser $\mathcal{G}$ and apply a cross-attention mechanism to integrate scene conditions $\mb{S}$, \textit{i.e.}, textual descriptions, road sketches, and 3D boxes.
Denoting $z^{b}_\tau(\epsilon)=\sqrt{\bar{\alpha}_\tau}z^{b}_0 + \sqrt{1-\bar{\alpha}_\tau}\epsilon$ as noisy latent, where $\tau$ is a timestep, $\epsilon\sim{\mathcal{N}(0, I)}$ is Gaussian noise, $\bar{\alpha}_\tau$ is hyper-parameter, the diffusion process is:
\begin{equation}
    \mathcal{L}_\text{diff} =
    \mathbb{E}_{{z^{b}_0},\epsilon,\tau, \mb{S}}\left[ \Vert
    \epsilon - \mathcal{G}^{b}_\theta(z^{b}_\tau(\epsilon),\tau, \mb{S})
    \Vert^2_2 \right].
\label{eq:diff}
\end{equation}
Specifically, as illustrated in \cref{fig:framework}, we incorporate the ControlNet~\cite{zhang2023adding} branch into the DiT model to enable road sketch conditions.
Inspired by PixArt-$\delta$~\cite{chen2024pixart}, we create a trainable copy of the first 13 blocks of the model. These duplicated blocks are integrated with the corresponding base blocks through a learnable zero linear layer. 
Each duplicated block combines the road sketch features, ensuring precise control with the provided sketch conditions.


\begin{figure}[t]
    \centering
    \includegraphics[width=\linewidth]{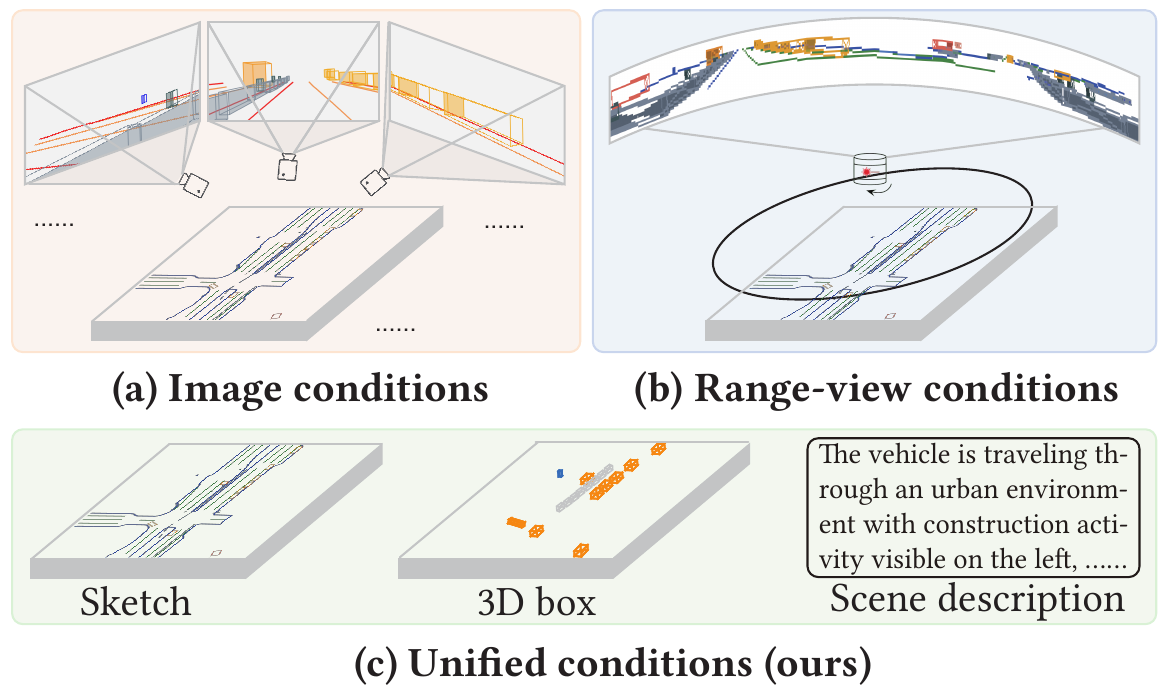}
    \caption{\textbf{Previous modality-specific conditions and our unified conditions.} }
    \label{fig:cindition}
    \vspace{-10pt}
\end{figure}
\mypara{Scene Conditions Encoding.}
To describe a driving scenario, we adopt comprehensive scene conditions outlined in \cite{gao2023magicdrive, ma2024unleashing}. 
As illustrated in \cref{fig:cindition}, unlike previous approaches that rely on modality-specific conditions, our method adopts unified conditions to generate aligned multimodal sensor data.
Specifically, scene conditions \(\mb{S}=\{\mb{M}, \mb{B}, \mb{T}\}\) include a road sketch \(\mb{M}\in\{0, 1\}^{w\times h\times c}\) representing a \(w\times h\) meter road area in BEV with \(c\) semantic classes, 3D bounding boxes \(\mb{B}=\{(\mb{b}_i, \mb{h}_i, \mb{l}_i, \mb{c}_i)\}_{i=1}^{n}\) where each object is described by a box \(\mb{b}_i=\{(x_j, y_j, z_j)\}_{j=1}^{8}\in\mathbb{R}^{8\times 3}\), heading $\mathbf{h}_i\in{[-180, 180)}^{n \times 1}$, instance id $\mathbf{l}_i\in{[0,1]}^{n \times 1}$, and caption \(\mathbf{c}_i\in\mathcal{C}\), textual descriptions \(\mb{T}\) summarying information for the whole scene (\textit{e.g.}, weather and time of day).
The layout entries, \textit{i.e.}, instance details such as box coordinates, heading, and ID, are encoded by the Fourier Embedder~\cite{mildenhall2021nerf}, $F$, and then are concatenated and processed through an MLP into a unified embedding.
The textual input is encoded into 200 tokens using the T5 \cite{raffel2020exploring} language model, $E_{\text{T5}}$.
The road sketches are extracted latent features by a pre-trained VAE, $E_{\text{VAE}}$. 
The encoding of unified scene conditions can be formulated as:
\begin{equation}
    \begin{aligned}
        & {\mathbf{B'}} = \text{MLP}( F({\mathbf{b} }) +  F({\mathbf{h}}) + F({\mathbf{l}}) + E_{\text{T5}}({\mathbf{c}}) ), \\[1mm]
        & ~~~~~ {\mathbf{M'}} = E_{\text{VAE}}( \mathbf{M} ), 
        ~~~~~~ {\mathbf{T'}} = E_{\text{T5}}( \mathbf{T} ).
    \end{aligned}
\label{eq:encoding}
\end{equation}
We incorporate these condition embeddings into the DiT model through cross-attention mechanisms, facilitating flexible and fine-grained control:
\begin{equation}
    \begin{aligned}
    {\mathbf{q}} = \text{MLP} ({\mathbf{z}_{\text{in}}^{b}}),~~
    {\mathbf{k}} &= \text{MLP} ({[\mathbf{B'}, \mathbf{T'}]}), ~~
    {\mathbf{v}} = \text{MLP} ({[\mathbf{B'}, \mathbf{T'}]}), \\
     \text{CA}(\mathbf{q}, \mathbf{k}, \mathbf{v}) &= \text{Softmax}(\frac{\mathbf{q} \cdot \mathbf{k}^{T}}{\sqrt{d}})\cdot \mathbf{v}.
    \end{aligned}
\label{equ:text_cross_attention}
\end{equation}
The road sketches feature is integrated in the duplicated blocks as:
\begin{equation}
    \begin{aligned}
        \mathbf{z}_{\text{out}}^b = \text{DiT}(\mathbf{z}_{\text{in}}^b) + \verb'Zero'(\text{Control}(\mathbf{z}_{\text{in}}^b + \mathbf{M'})),
    \end{aligned}
\label{eq:STDiT}
\end{equation}
where $\verb'Zero'$ denotes the learnable zero linear layer.

\mypara{Optimization.}
With the latest advancement on LDMs~\cite{opensora}, we replace IDDPM~\cite{nichol2021improved} with rectified flow~\cite{liuflow} for increased stability and reduced inference steps. 
Rectified flow defines the forward process between data and normal distributions as $z^{b}_\tau=(1-\tau)z^b_0+ \tau z^b_1$, and the loss function in \cref{eq:diff} is rewritten as:
\begin{equation}
 \mathcal{L}_\text{rf} = \mathbb{E}_{z^b_1,z^b_0,\tau, \mb{S}} \left[ \Vert \mathcal{G}_{\theta}^b(z^{b}_\tau, \tau, \mathbf{S})-(z^b_1-z^b_0)   \Vert^2_2 \right].
\end{equation}

Moreover, we extend the Classifier-free Guidance (CFG)~\cite{ho2022cfg} strategy from the text condition to 3D boxes and road sketches to enhance control precision and visual quality.
CFG aims to enhance the alignment between generated images and specified conditions, which simultaneously performs both conditional and unconditional denoising during training and combines the two estimated scores during inference. 
In practice, we randomly set each condition to a null $\phi$ with a 5\% probability during training. The guidance scale $\lambda_M, \lambda_B, \lambda_T$ controls the alignment between the sampling results and the conditions.
Drawing inspiration from IP2P~\cite{brooks2023instructpix2pix}, we apply the unconditional denoising results to each condition individually, which can be formulated as:
\begin{equation}
    \begin{aligned}
        \tilde{\mathcal{G}_\theta}( &z^{b}_\tau ,B ,M , T) = \mathcal{G}_\theta(z^{b}_\tau, \phi, \phi, \phi) \\
        &+\lambda_T\cdot(\mathcal{G}_\theta(z^{b}_\tau, \phi, \phi, T)-\mathcal{G}_\theta(z^{b}_\tau,\phi,\phi,\phi)) \\
        &+\lambda_M\cdot(\mathcal{G}_\theta(z^{b}_\tau, \phi, M, T)-\mathcal{G}_\theta(z^{b}_\tau, \phi,\phi,T)) \\
        &+\lambda_B\cdot(\mathcal{G}_\theta(z^{b}_\tau, B, M, T)-\mathcal{G}_\theta(z^{b}_\tau, \phi, M, T)).
    \end{aligned}
\label{eq:cfg}
\end{equation}


\section{Experiment}
\label{sec:exp}
Our experiments are conducted on the popular NuScenes dataset~\cite{caesar2020_nuscenes}. For additional details, including dataset, metrics, baselines, and implementations, please see \cref{sec:details} in the supplementary material.

\subsection{Main Results}
\label{subsec:mainres}
\begin{table}[t]
  \centering 
  \caption{\textbf{Multimodal sensor generation results.}}
  \addtolength{\tabcolsep}{-0.4pt}
    \begin{tabularx}{\linewidth}{l|c|ccc}
  \toprule
 \multirow{2}{*}{Method}  & \multirow{2}{*}{Tokenizer}  & \multicolumn{3}{c}{\textit{\textbf{Camera Generation}}}   \\
      \cmidrule(r){3-5} & & FID$\downarrow$ & CLIP$\uparrow$  & mAP$\uparrow$  \\
      \midrule    

        \multicolumn{3}{l}{\textit{\textbf{Single-Modality}}}    \\
        \midrule

BEVGen~\cite{swerdlow2023_bevgen}    & VQ-VAE      &25.54                     &71.23           & --                                  \\
BEVControl~\cite{yang2023bevcontrol}         & VQ-VAE          &24.85                      &82.70 & 19.64                                                  \\
DriveDreamer~\cite{wang2023drivedreamer}      & VQ-VAE           &  52.60                 &--              &--                                    \\
DriveDreamer-2 ~\cite{zhao2024drivedreamer-2}      & VQ-VAE           &  25.00                &--              &-- \\
WoVoGen~\cite{lu2023wovogen}& VQ-VAE           &  27.60 &--              &--       \\
MagicDrive~\cite{gao2023magicdrive}        & VQ-VAE       &16.20                      & 82.47 & 12.30                                                \\                                
Panacea~\cite{wen2023panacea}          & VQ-VAE           &16.96            & 84.23          &--                                                  \\
Drive-WM~\cite{wang2023driveWM}       & VQ-VAE              &15.80                    &-- & 20.66                                                 \\
MagicDriveDiT~\cite{gao2024magicdrivedit} & VQ-VAE  & 20.91 & -- & 17.65 \\
 \rowcolor{mygray}  \bf\name{}   & \bf UAE-C & 22.15 &  82.76 &  19.57     \\

    \midrule
        \multicolumn{3}{l}{\textit{\textbf{Unified}}}    \\
        \midrule
        
 \rowcolor{mygray}  \bf\name{}   & \bf UAE-LC  & 21.01 & 83.54 &  20.41 \\
                \midrule

      \midrule    
 \multirow{2}{*}{Method}  & \multirow{2}{*}{Tokenizer}  & \multicolumn{3}{c}{\textit{\textbf{LiDAR Generation}}}  \\
      \cmidrule(r){3-5} & & FRD$\downarrow$     & MMD$\downarrow$  & JSD$\downarrow$     \\ 
      \midrule    

        \multicolumn{3}{l}{\textit{\textbf{Single-Modality}}}    \\
        \midrule

LiDARVAE~\cite{caccia2019deep}   & 2DGrid-VAE & --& 11.0& - \\
LiDARGen~\cite{zyrianov2022learning}   & N/A & --     & 19.0        & 0.160     \\
RangeLDM~\cite{hu2024rangeldm} & Range-VAE & 492.49  &  2.75  & {0.054} \\
LidarDM~\cite{zyrianov2024lidardm}   & SDF-VAE &-- & 3.51 &  0.118\\
 \rowcolor{mygray}  \bf\name{}   & \bf UAE-L  & 562.89  & 3.17 & 0.117 \\

     \midrule
        \multicolumn{3}{l}{\textit{\textbf{Unified}}}    \\
        \midrule
        
 \rowcolor{mygray}  \bf\name{}   & \bf UAE-LC  & 519.73 & 2.94 & 0.105 \\

\bottomrule

      \multicolumn{5}{l}{\footnotesize MMD has been multiplied by \({10^{4}}.\) }\\
      \multicolumn{5}{l}{\footnotesize "LC" represents the LiDAR and Camera fusion. }\\
    \end{tabularx}
        \vspace{-15pt}
  \label{tab:mainres}
\end{table}

\begin{table}[t]
  \centering 
  \caption{\textbf{Generalizable multimodal reconstruction results.}}
  \addtolength{\tabcolsep}{-0.3pt}
    \begin{tabularx}{1\linewidth}{l|cc|cc}
  \toprule
 \multirow{2}{*}{\textit{\textbf{Camera Recon}}} & \multicolumn{2}{c|}{Train set} & \multicolumn{2}{c}{Val set} \\
      \cmidrule(r){2-5} & PSNR$\uparrow$ & SSIM$\uparrow$  &PSNR$\uparrow$ & SSIM$\uparrow$ \\ 
      
\midrule

SelfOcc~\cite{huang2024selfocc} & 20.67 & 0.556 & --&--      \\
UniPAD~\cite{hu2023uniad} &   19.44 & 0.497 & --&--      \\
DistillNeRF~\cite{wang2024distillnerf}  & 28.01 & 0.872 & --&--      \\        
 \rowcolor{mygray}  \bf UAE-C   & 30.29 & 0.908  & 30.13 & 0.903   \\
 \rowcolor{mygray}  \bf UAE-LC & \bf 30.45 & \bf 0.913 & \bf 30.21 & \bf 0.909  \\

\midrule
\midrule
 \multirow{2}{*}{\textit{\textbf{LiDAR Recon}}} & \multicolumn{2}{c|}{Train set} & \multicolumn{2}{c}{Val set}  \\
      \cmidrule(r){2-5}  & Chamfer$\downarrow$ & F-score$\uparrow$ &  Chamfer$\downarrow$ & F-score$\uparrow$\\ 

      \midrule
      
 \rowcolor{mygray}  \bf UAE-L    & 0.869 & 0.734 & 1.068& 0.713 \\
 \rowcolor{mygray}  \bf UAE-LC    & \bf 0.634 & \bf 0.763   &\bf 0.793  & \bf 0.742 \\

      \bottomrule

    \end{tabularx}
  \label{tab:vae}
  \vspace{-5pt}
\end{table}

\begin{table}[ht]
\centering
\caption{Generation data augmentation for perception.}
\label{tab:bevfusion}
\addtolength{\tabcolsep}{0pt}
 \begin{tabularx}{0.85\linewidth}{l|c|cc}
\toprule
Method & Modality & mAP$\uparrow$ & NDS$\uparrow$ \\
\midrule

BEVFormer~\cite{li2022bevformer} &C & 25.2 & 35.4  \\
\rowcolor{mygray} + \textbf{\name{}} & C & \textbf{27.1} {\scalebox{0.92}[0.92]{\textbf{(+1.9)}}}& \textbf{37.1} {\scalebox{0.92}[0.92]{\textbf{(+1.7)}}}\\
\midrule


BEVFusion~\cite{liu2023bevfusion}            & LC     & 68.5 & 71.4    \\
\rowcolor{mygray} + \textbf{\name{}} & LC & \textbf{70.1} {\scalebox{0.92}[0.92]{\textbf{(+1.6)}}} & \textbf{72.8} {\scalebox{0.92}[0.92]{\textbf{(+1.4)}}} \\

\bottomrule
\end{tabularx}

\end{table}






\begin{table}[ht]
\centering
\caption{Generation data augmentation for planning.
}

\label{tab:uniad}

\addtolength{\tabcolsep}{-2.5pt}
 \begin{tabularx}{0.95\linewidth}{l|c|c|c}
\toprule

Method & Modality &{Avg. L2 (m) $\downarrow$} & 
{Avg. Collision (\%) $\downarrow$} \\
\midrule

UniAD~\cite{hu2023uniad} &C & 1.03 & 0.31 \\
\rowcolor{mygray} + \textbf{\name{}} & C & \textbf{0.99} {\scalebox{0.92}[0.92]{\textbf{(+3.9\%)}}}& \textbf{0.29} {\scalebox{0.92}[0.92]{\textbf{(+6.4\%)}}}\\
\midrule
FusionAD~\cite{ye2023fusionad} & LC &0.81 & 0.12 \\
\rowcolor{mygray} + \textbf{\name{}} & LC & \textbf{0.77} {\scalebox{0.92}[0.92]{\textbf{(+4.9\%)}}}& \textbf{0.11} {\scalebox{0.92}[0.92]{\textbf{(+8.3\%)}}}\\
\bottomrule
\end{tabularx}

\end{table}

\mypara{Unified multimodal sensor generation.}
As shown in \cref{tab:mainres}, we compare our model with specialized single-modality generation methods. Although our method does not achieve state-of-the-art (SOTA) performance for every metric, \name{}, as a unified multimodal framework, achieves comparable or even superior quality in camera and LiDAR sensor generation. Specifically, \name{} achieves 21.01 FID and 20.41 mAP for generated camera data, and 2.94 × $10^{-4}$ MMD and 0.105 JSD for LiDAR data. 
Furthermore, thanks to our independent BEV encoders for each modality, our \UAE{} (\textit{i.e.}, \UAE{}-LC) also supports single-modality operation, \textit{i.e.}, \UAE{}-C and \UAE{}-L. Experimental results show that multimodal generation consistently outperforms our single-modality generation, further demonstrating the effectiveness of our unified model.
We hope our framework can inspire the community and foster collaborative efforts to improve the results to SOTA together.

\mypara{Generalizable LiDAR-camera reconstruction.}
As shown in \cref{tab:vae}, our \UAE{} achieves state-of-the-art performance in generalizable LiDAR-camera multimodal reconstruction. Specifically, \UAE{} significantly outperforms the previous best generalizable single-camera modality method, DistillNeRF~\cite{wang2024distillnerf}, with a notable improvement (\textit{i.e.},  +2.44 PSNR). 
Moreover, our \UAE{}-L and \UAE{}-LC first achieves generalizable LiDAR reconstruction with the chamfer distance of 0.869 and 0.634, respectively.
Additionally, multimodal reconstruction also surpasses single-modality reconstruction, further highlighting the advantages of our unified framework.

\mypara{Downstream tasks.}
As shown in \cref{tab:bevfusion} and \cref{tab:uniad}, we utilize our \name{} to produce augmented data with corresponding conditions, aiming to enhance downstream tasks.
As illustrated by the improvements, our \name{} effectively generates multimodal sensor data, facilitating enhanced perception and planning in autonomous driving. Specifically, \name{} boosts multimodal BEVFusion with +1.6 mAP and FusionAD with +4.9\% L2 metric. 

\subsection{Qualitative Results}
\label{subsec:vis}
\begin{figure}[ht]
    \centering
    \includegraphics[width=\linewidth]{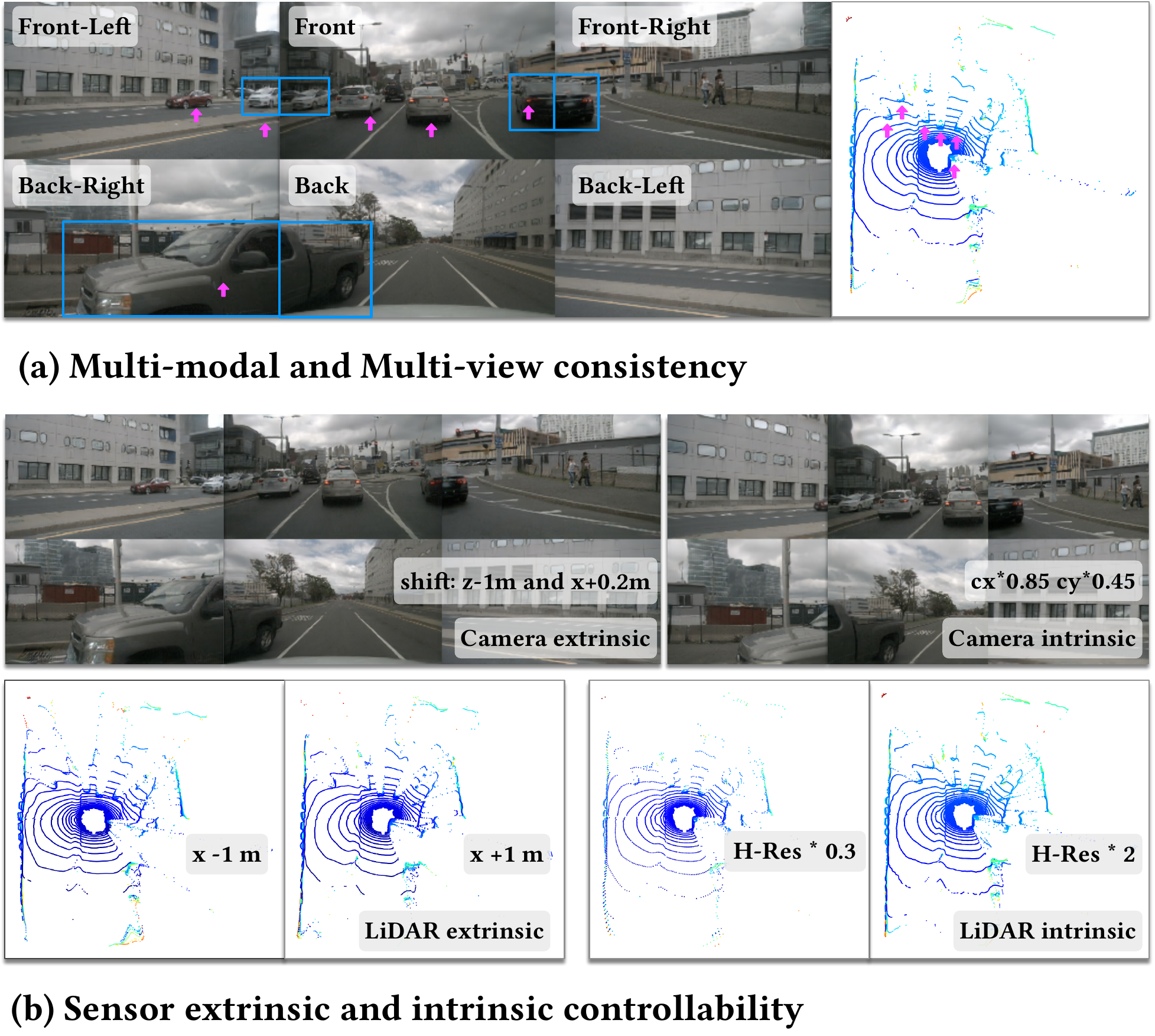}
    \caption{\textbf{Qualitative results for multimodal sensor generation.} H-res denotes the horizontal resolution.} 
    \label{fig:vis}
\end{figure}
\mypara{Multi-modal and multi-view consistency.}
As shown in \cref{fig:vis} (a), our \name{} exhibits excellent multi-modal and multi-view consistency, as highlighted by the blue boxes and pink arrows. This demonstrates the effectiveness of our unified method, which leverages unified BEV features as a global scene constraint.

\mypara{Sensor intrinsic and extrinsic controllability.}
As shown in \cref{fig:vis} (b), UAE enables the flexible adjustment of sensor parameters, including both intrinsic and extrinsic settings, during the rendering process, which showcases the superior design of our framework.

\subsection{Ablation Study}
\label{subsec:ablation}

\begin{table}[t]
  \centering 
  \caption{\textbf{Ablation on UAE.}}
  \addtolength{\tabcolsep}{2.2pt}
    \begin{tabularx}{0.9\linewidth}{l|ccc}
  \toprule

 Module& PSNR$\uparrow$ & SSIM$\uparrow$ & LPIPS$\downarrow$\\
      \midrule    



\multicolumn{3}{l}{\textit{\textbf{Architecture}}}    \\
\midrule
A1-MLP {\scalebox{0.8}[0.8]{\textcolor{brown}{Fig5 (b)}}}  & 29.42 &0.878    & 0.049 \\ 
A2-w/o-decoder {\scalebox{0.8}[0.8]{\textcolor{brown}{Fig5 (c)}}} & 26.61 & 0.802  & 0.212  \\
\rowcolor{mygray} UAE {\scalebox{0.8}[0.8]{\textcolor{brown}{Fig5 (a)}}} & \textbf{30.21} & \textbf{0.909} & \textbf{0.033}  \\





\midrule
\multicolumn{3}{l}{\textit{\textbf{Render Representation}}}    \\
\midrule

NeRF {\scalebox{0.8}[0.8]{\textcolor{brown}{Fig5 (d)}}}  & 29.97 & 0.891     & 0.069   \\
\rowcolor{mygray} SDF  {\scalebox{0.8}[0.8]{\textcolor{brown}{Fig5 (a)}}} & \textbf{30.21} &  \textbf{0.909} & \textbf{0.033}      \\

      \bottomrule

    \end{tabularx}
  \label{tab:uae_ab}
    \vspace{-5pt}
\end{table}

\begin{figure}[t]
    \centering
    \includegraphics[width=\linewidth]{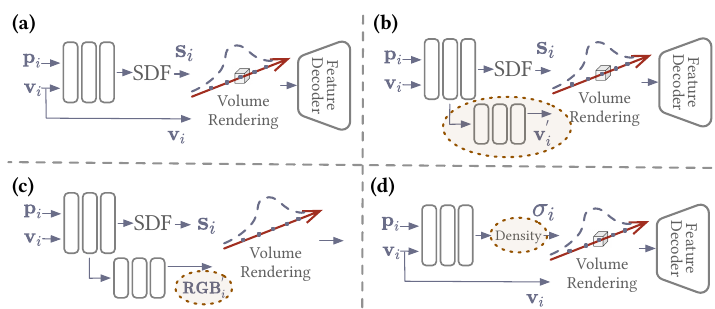}
    \caption{\textbf{Different architecture designs of UAE.} }
    \label{fig:vae_ablation}
\end{figure}

\mypara{Ablations on UAE.}
As shown in \cref{tab:uae_ab} and \cref{fig:vae_ablation}, we investigate various designs of the render decoder in \UAE{}. For A1-MLP, we introduce additional MLPs to further extract voxel features; however, this does not improve the results and instead increases the number of parameters. For A2-w/o-decoder, we use additional MLPs to directly output RGB values, removing the feature decoder altogether. This leads to a significant performance drop, highlighting the importance of the feature decoder. 
Additionally, we ablate different representations for rendering, including NeRF and SDF, both of which produce comparable results. The SDF performs slightly better and is therefore adopted in the final design.

\section{Discussion}
In this section, we aim to provide valuable insights for the unified multimodal generation field. More discussions about Future Work of Multimodal Generation, Limitations and Unsuccessful Attempts, please see \cref{sec:additional_discussion} in the supplementary material.

\subsection{Mutimodal Sensor Generation Solutions}
\begin{figure}[t]
    \centering
    \includegraphics[width=\linewidth]{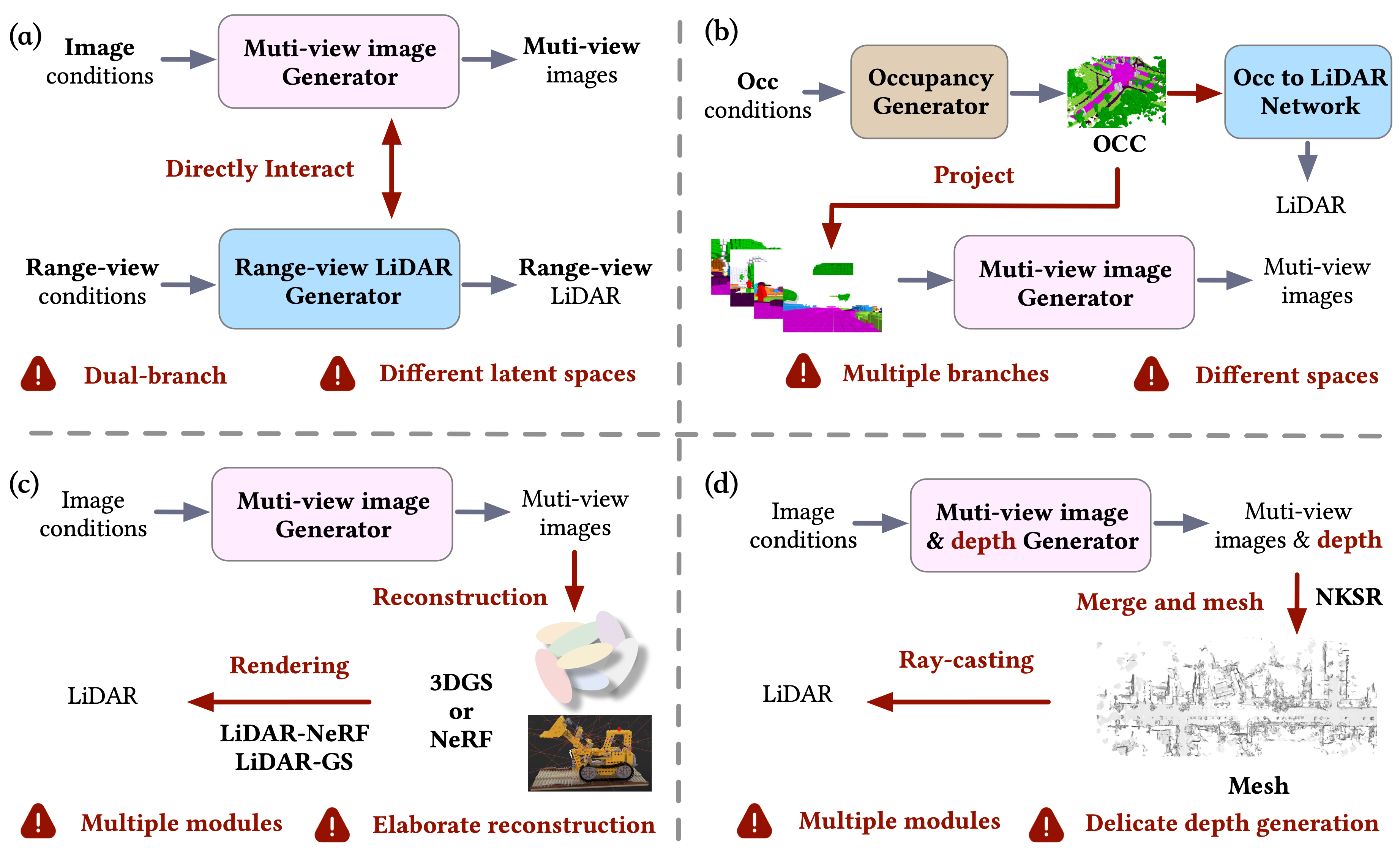}
    \caption{\textbf{Multimodal sensor generation solutions (zoom-in for better views).}}
    \label{fig:other}
    \vspace{-6pt}
\end{figure}
During the early design phase of \name{}, we explored various alternative frameworks. Although these ideas were not pursued further due to certain limitations, \textit{e.g.}, not unified, we present them here to share potential ideas for future research.
As shown in \cref{fig:other}, these frameworks build upon the camera generation pipeline as the primary branch, leveraging its superior camera generation performance, while incorporating the LiDAR modality in different ways.

\begin{itemize}[leftmargin=*]
    \item a) A dual-branch design, where each modality operates in its own feature space and cross-modal alignment is enforced by attention mechanisms. Due to the inherent differences in representation spaces, achieving alignment proves challenging, which leads to limited performance potential for the overall framework.

    \item b) Some studies~\cite{yan2024drivingsphere, li2024syntheocc, lu2024infinicube} use semantic occupancy as an intermediate representation to improve generation quality. Based on this approach, we explored adding an occupancy-to-LiDAR branch. However, this results in a complex framework with multiple branches, and its performance is restricted by limitations in the occupancy generation process, such as resolution.

    \item c) Expanding upon the existing camera branch, we incorporated a reconstruction stage using NeRF or 3DGS to reconstruct the scene, followed by rendering LiDAR data through methods such as~\cite{tao2023lidar,chen2024lidar}. However, this framework suffers from inefficiencies due to its multi-stage nature, particularly the time-consuming and labor-intensive reconstruction process.

    \item d) Building on the existing camera branch, we integrated it with per-view depth map generation and then merged the generated depth maps into a mesh using methods such as NKSR~\cite{huang2023neural}. Subsequently, LiDAR data can be synthesized through ray-casting. However, this framework is limited by its multi-stage complexity and the challenges of generating precise depth.

\end{itemize}


\section{Conclusion}
In this paper, we introduced \name{}, a unified multimodal sensor generation framework for autonomous driving that enables the unified generation of aligned LiDAR and camera data. 
Our approach addresses the limitations of existing single-modality generation methods by establishing a unified BEV-based representation space, proposing \UAE{}, a generalizable multimodal reconstruction for multimodal autoencoder, and incorporating a ControlNet-Transformer model to synthesize multimodal sensor data under flexible conditions.
Extensive experiments demonstrate that our framework not only achieves state-of-the-art performance in multimodal reconstruction but also generates LiDAR and camera data with cross-modality alignment and flexible sensor control. 
These advancements enhance the quality and usability of synthetic sensor data, further benefiting downstream tasks such as perception and planning in autonomous driving.
Moreover, we provide valuable insights into the designs of the unified multimodal generation framework, hoping to inspire future research 
into more efficient multimodal generation.

\section{Acknowledgments}
This work is supported by National Key Research and Development Program of China (2024YFE0203100), National Natural Science Foundation of China (NSFC) under Grants No.62476293, Shenzhen Science and Technology Program No.GJHZ20220913142600001, Nansha Key R\&D Program under Grant No.2022ZD014, and General Embodied AI Center of Sun Yat-sen University.

{
    \small
    \bibliographystyle{ieeenat_fullname}
    \bibliography{main}
}
\clearpage
\newpage


\section{Additional Discussion}
\label{sec:additional_discussion}

\subsection{Future Work of Multimodal Generation}
\begin{figure}[h]
    \centering
    \includegraphics[width=\linewidth]{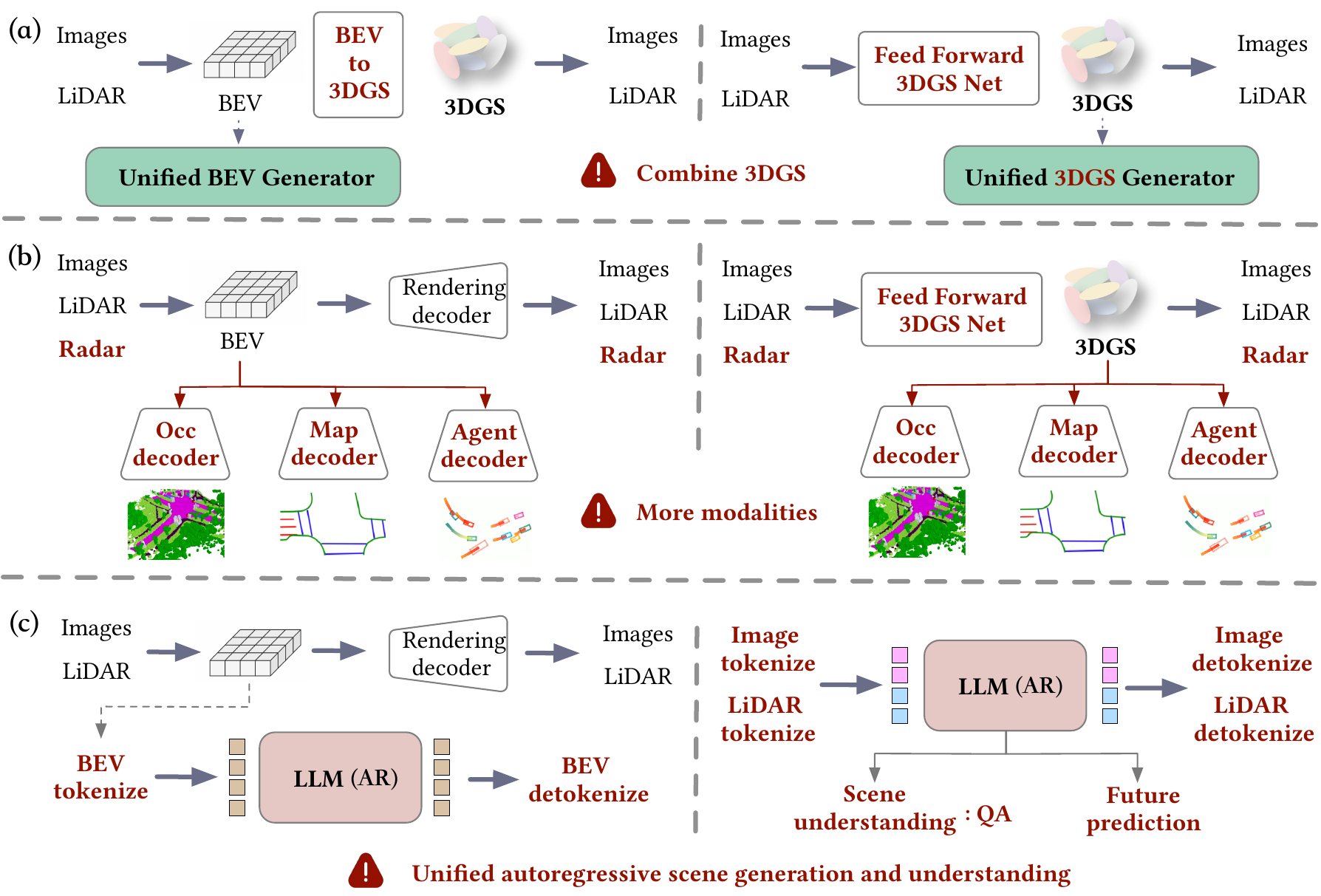}
    \caption{\textbf{Unified multimodal generation (zoom-in for better views).}}
    \label{fig:future}
      \vspace{-6pt}
\end{figure}
Thanks to our unified framework, as illustrated in \cref{fig:future}, there remain many promising directions to explore:
\begin{itemize}[leftmargin=*]
    \item a) In this paper, we employ NeRF/SDF-based rendering techniques. However, our approach can be readily extended to incorporate more advanced rendering methods, \textit{i.e.}, 3DGS. For instance, as illustrated on the left, a BEV-to-3DGS module can be integrated as~\cite{xu2024gaussianpretrain, zhang2024visionpad}. Alternatively, as shown on the right, a feed-forward 3DGS design can be adopted as~\cite{tian2024drivingforward, chen2024g3r} and then followed by a 3DGS generator such as~\cite{zhou2024diffgs, lin2025diffsplat}.

    \item b) Our unified BEV space exhibits significant scalability, enabling straightforward extensions to additional sensors like radar. Additionally, it can be adapted to other tasks previously supported by BEV representations, such as the prediction or generation of occupancy~\cite{wang2023openoccupancy}, HD map~\cite{dong2024superfusion}, or agent trajectory~\cite {li2025end}. These tasks can be accomplished by simply attaching the corresponding decoder to either the BEV or 3DGS representation. Incorporating multiple BEV-supported tasks can enrich and strengthen feature robustness, but it may also increase complexity, introducing challenges such as task alignment and modality interference.

    \item c) Currently, our approach employs a diffusion model for generation. Inspired by autoregressive models such as Emu3~\cite{wang2024emu3}, we could also leverage large language models (LLMs) for unified generation. This could be achieved by reusing our UAE module or directly tokenizing multimodal sensor inputs for interaction.
    The advantage of employing LLMs lies in their ability to further unify scene generation and understanding as~\cite{zhou2025hermes, wang2025mmgen}, enabling support for tasks such as question answering (QA).

\end{itemize}

\subsection{Limitations}
This paper proposes a generalizable LiDAR-camera reconstruction method and a unified multimodal sensor generation framework. 
Given the extensive scope of this work, there remain several aspects that are not yet fully developed:

\begin{itemize}[leftmargin=*]
    \item The current generation quality does not yet reach state-of-the-art performance, particularly for the camera modality. We hypothesize that this is due to the per-pixel ray sampling rendering strategy, which lacks coherence between sampled pixels. This leads to weak correlations between BEV features. Incorporating spatial regularization techniques, such as Total Variation (TV) loss as voxel-based NeRFs~\cite{fridovich2022plenoxels}, may further enhance performance. 
    The observed trade-offs may also arise from the limited performance of the adopted generative model, i.e., Open-Sora. \cref{fig:future} also outlines future directions—such as integrating 3DGS or using AR models—that may enhance performance. Overall, we believe a unified framework remains a promising path and hope to inspire the community to push unified models beyond the specialized single-modality generators collaboratively.    
    \item In this paper, we only present single-frame generation. However, our codebase inherently supports video generation with temporal attention. With some engineering for training and inference, it can be extended to support video synthesis, as concurrent works like MagicDriveDiT~\cite{gao2024magicdrivedit}.
    \item Our \UAE{} currently only utilizes six images from a single frame, which limits the quality of wide-angle sensor extrinsic adjustments. Future work can explore the integration of temporal information from adjacent frames.
\end{itemize}

\subsection{Unsuccessful Attempts}
In the early stages of developing our framework, we also encountered failures and setbacks along the way. We share our experiences here to provide insights, but this does not imply that these attempts are inherently ineffective.

\begin{itemize}[leftmargin=*]
    \item Avoid initializing rendering models with pretrained perception models. Some prior works~\cite{yang2024unipad, xu2023nerf} have explored using rendering-based pretraining for perception models. However, we found that the reverse—using a pretrained perception model to initialize the rendering model—led to poor results, even when only partial components like the LSS projection were loaded. This is likely because rendering tasks require low-level features, whereas perception models operate on high-level semantics. This finding aligns with discussions in NeRFDet~\cite{xu2023nerf}.
    
    \item Camera-only rendering tends to overfit in BEV features. When using only the camera modality, the model tends to overfit to uninformative BEV feature maps with no discriminative structure. This issue may arise from the lack of supervision in the LSS projection process. Interestingly, when vector quantization is incorporated, this overfitting is mitigated, and the BEV features begin to exhibit meaningful road structure.

    \item LiDAR modality dominates in joint training. During joint training of LiDAR and camera modalities, LiDAR tends to learn more easily, possibly due to the inherently structured voxel-based representation. While camera features are projected via LSS, which may be less efficient. Additionally, the camera RGB has a higher channel than LiDAR depth, suggesting that camera loss should be up-weighted in joint training to ensure balanced learning.

    \item Shared MLP for both modalities leads to performance degradation. Sharing a single MLP between LiDAR and camera modalities did not result in positive cross-modal benefits; in fact, it reduced performance. As a result, we adopted separate MLPs for each modality. Similar observations were also analyzed in AlignMiF~\cite{tao2024alignmif}.

    \item Patch-wise ray sampling with a feature decoder improves convergence over random sampling. During rendering, we observed that sampling rays in patches and feeding them into the decoder led to faster and more stable convergence and better rendering quality compared to random sampling strategies.

    \item High-resolution BEV features slow down DiT training. Generating BEV features at higher resolutions (\textit{e.g.}, 180×180) leads to much slower convergence compared to lower resolutions (\textit{e.g.}, 90×90).
    
\end{itemize}

\section{Additional Details}
\label{sec:details}

\mypara{Datasets.}
We evaluate our method on the large-scale nuScenes~\cite{caesar2020_nuscenes} dataset, which comprises multimodal data such as multi-view images from 6 cameras and point clouds from LiDAR, scene descriptions, and annotations about the boxes and map. 
It contains 700 scenes for training and 150 scenes for validation, characterized by various street driving scenarios. 
Each scene includes 20 seconds at a frame rate of 12Hz and annotations at a frame of 2Hz keyframe.

\mypara{Evaluation metrics.}
1) For generalizable LiDAR-Camera reconstruction, \textit{i.e.}, \UAE{}, we report peak signal-to-noise ratio (PSNR) and structural similarity index measure (SSIM)~\cite{wang2004ssim}, and the learned perceptual image patch similarity (LPIPS)~\cite{zhang2018lpips} 
for camera synthesis, and report Chamfer Distance (C-D) and F-Score for LiDAR synthesis.
2) For multimodal sensor generation, we report Frechet Inception Distance (FID)~\cite{heusel2017_fid}, CLIP score~\cite{yang2023bevcontrol} and 3D detection mAP (evaluate by BEVFormer~\cite{li2022bevformer}) for generated camera data evaluation, and report Frechet point cloud distance (FPD), Jensen–Shannon divergence (JSD), and minimum matching distance (MMD) for generated LiDAR data evaluation.
3) For downstream tasks, we follow the official evaluation and report nuScenes Detection Score (NDS) and mean Average Prediction (mAP) for 3D object detection, and report L2 error and collision rate for planning.

\begin{figure}[ht]
    \centering
    \includegraphics[width=\linewidth]{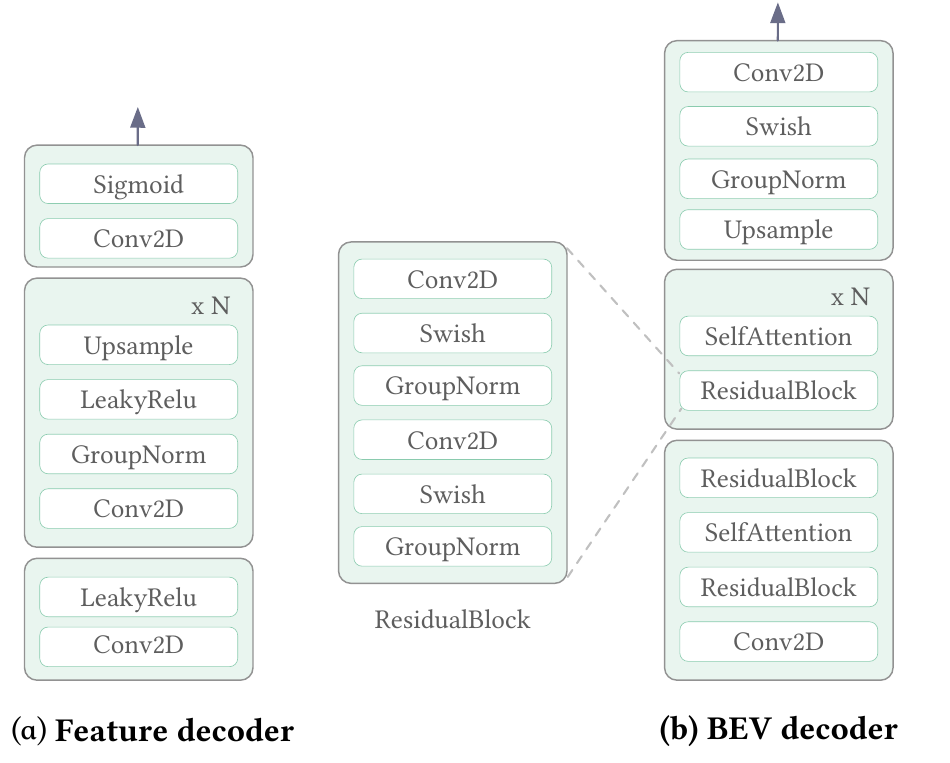}
    \caption{\textbf{Detailed architectures of the feature decoder and BEV decoder in \UAE{}.} }
    \label{fig:UAEdecoder}
\end{figure}
\mypara{Baselines.}
1) For generalizable LiDAR-camera reconstruction, \textit{i.e.}, \UAE{}, since there are no generalizable LiDAR or multimodal methods, we compare our model with SOTA generalizable camera reconstruction methods, \textit{i.e.}, SelfOcc~\cite{huang2024selfocc}, UniPAD~\cite{hu2023uniad}, and DistillNeRF~\cite{wang2024distillnerf}.
2) For multimodal sensor generation, we compare our model with previous single-modality generation methods. For camera sensor generation, we compare with SOTA camera 
generation approaches, MagicDrive~\cite{gao2023magicdrive}, DriveDreamer~\cite{wang2023drivedreamer}, and so on. For LiDAR sensor generation, we compare with SOTA LiDAR generation approaches, 
LiDARGen~\cite{zyrianov2022learning}, RangeLDM~\cite{hu2024rangeldm} and LidarDM~\cite{zyrianov2024lidardm}.
3) For downstream tasks, we evaluate our model on popular BEV detection models (BEVFusion~\cite{liu2023bevfusion} and BEVFormer~\cite{li2022bevformer}) and planning models (UniAD~\cite{hu2023uniad} and FusionAD~\cite{ye2023fusionad}) to analyze the sim-to-real gap between the generated data and real-world data.

\mypara{Implementation Details.}
1) For generalizable LiDAR-camera reconstruction, \textit{i.e.}, \UAE{}, our code is base on UniPAD~\cite{hu2023uniad}. We employ ConvNeXt-s~\cite{liu2022convnext} and VoxelNet~\cite{yan2018second} as the default camera and LiDAR encoders, respectively.
The input image size is $1600 \times 900$, and the voxelization size is $[0.075, 0.075, 0.2]$.
The unified voxel representation with the size of $[0.6, 0.6, 1.6]$ and shape of $180 \times 180 \times 5$ is constructed across modalities.
For the vector quantization, following~\cite{zhang2024vq}, we use the BEV patch size $P=2$ and codebook size of $K=512$ with a dimension of $E=8$. For the patch embedder, we employ two 3$\times$3 convolutional modules, with channels of $64, 128$.
For the BEV decoder, the detail structure is shown in \cref{fig:UAEdecoder} (b). We also use the exponential moving average (EMA) with decay of 0.99 and epsilon value of 1e-5 to update the codebook embedding.
For rendering decoder, we utilize a $6$-layer MLP for SDF, the detailed feature decoder is shown in \cref{fig:UAEdecoder} (a).
In the rendering phase, $N=192$ points per ray are sampled.
For training, we train \UAE{} for $24$ epochs, using the AdamW optimizer, with an initial learning rate of $2e^{-4}$.
We also adopt a cosine learning rate decay schedule with warm-up in the first 500 steps.
2) For multimodal sensor generation, our code is based on OpenSora~\cite{opensora}. The model is optimized using the AdamW optimizer with a learning rate of $1e^{-4}$ for $30$k steps, incorporating $1.5$k steps warm-up. For inference, we sample using rectified flow with 30 steps, and set $\lambda_T$, $\lambda_M$, $\lambda_B$ to 1.0, 2.0, and 2.0, respectively.
All the models are trained on 8 NVIDIA A100 (80G) GPUs.

\section{Additional Results}

\begin{table}[t]
  \centering 
  \caption{\textbf{Ablation on the generation model.}}
  \addtolength{\tabcolsep}{1.7pt}
    \begin{tabularx}{0.9\linewidth}{l|cccc}
  \toprule

 Module& FID$\downarrow$ & mAP$\uparrow$ &MMD$\downarrow$ &JSD$\downarrow$\\
      \midrule    
w/o Sketch & 27.34 & 16.23 & 5.31 & 0.139 \\
w/o 3D boxes & 25.57 & 16.58 & 4.79 & 0.124 \\
w/o Text prompt & 23.56 & 17.67 & 3.49 & 0.117 \\
  \midrule 
\rowcolor{mygray} Full model & \textbf{21.01} & \textbf{20.41} & \textbf{2.94} & \textbf{0.105}\\

      \bottomrule
\multicolumn{5}{l}{\footnotesize MMD has been multiplied by \({10^{4}}.\) }\\
    \end{tabularx}
  \label{tab:sora}
  \vspace{-5pt}
\end{table}

\mypara{Ablations on the generation model.}
As shown in \cref{tab:sora}, we ablate different conditions of the generation model. The multiple conditions provide multi-level control signals, enabling the generation of diverse scenes and contributing to the improvement of both LiDAR and image quality. Among these conditions, the sketch has the greatest impact due to its strong correspondence with the BEV. Please see the BEV visualization in \cref{fig:bev} in the supplementary.

\begin{table}[ht]
\centering
\small
\caption{Computational comparison.}
\label{tab:comp}
\addtolength{\tabcolsep}{-1.5pt}
 \begin{tabularx}{\linewidth}{l|c|ccc}
\toprule
Method & Condition & Inference & Backbone & Decoder \\
\midrule

Camera &	Project to image&	4.1s/sample	&1.1B	&83M \\
LiDAR	& Project to range-view	&3.8s/sample	&46.1M &	12.7M \\
OmniGen	& No need	&5.2s/sample	&1.1B&	39.8M \\

\bottomrule
\end{tabularx}

\end{table}

\mypara{Computational comparison.}
We provide computational comparisons in \cref{tab:comp}. OmniGen adds only a slight increase in inference time compared to the camera pipeline while enabling unified multimodal sensor generation. Unlike prior camera models that generate six image features, our generator produces only a single BEV feature. Although decoding multimodal data introduces some overhead, overall inference time remains comparable. Current generative models still lack efficiency—adequate for offline generation but far from real-time performance needed for tasks like closed-loop simulation with VLA models. Encouragingly, recent advances in single-step diffusion models offer a promising path forward.

\begin{table}[ht]
\centering
\small
\caption{Multimodal consistency.}
\label{tab:con}
\addtolength{\tabcolsep}{-2.2pt}
 \begin{tabularx}{0.85\linewidth}{l|cc}
\toprule
Method  & mAP$\uparrow$ & NDS$\uparrow$ \\
\midrule

BEVFusion~\cite{liu2023bevfusion}               & 68.5 & 71.4    \\

+ MagicDrive~\cite{gao2023magicdrive} \& RangeLDM~\cite{hu2024rangeldm}	& 68.1(-0.4)	& 71.2(-0.2) \\

 + \name{} & 70.1(+1.6) &72.8(+1.4) \\

\bottomrule
\end{tabularx}

\end{table}

\mypara{Multimodal consistency.}
Since no direct metric exists for multimodal consistency, we use improvements in downstream multimodal tasks as an indirect quantitative proxy, and compare with popular baselines, MagicDrive~\cite{gao2023magicdrive} \& RangeLDM~\cite{hu2024rangeldm}, in \cref{tab:con}. These separate models operate in different representation spaces, causing cross-modal inconsistencies that degrade downstream detection performance. In contrast, our unified multimodal framework preserves consistency across modalities and shows clear advantages in downstream tasks.

\section{Additional Visualizations}
\begin{figure}[t]
    \centering
    \includegraphics[width=\linewidth]{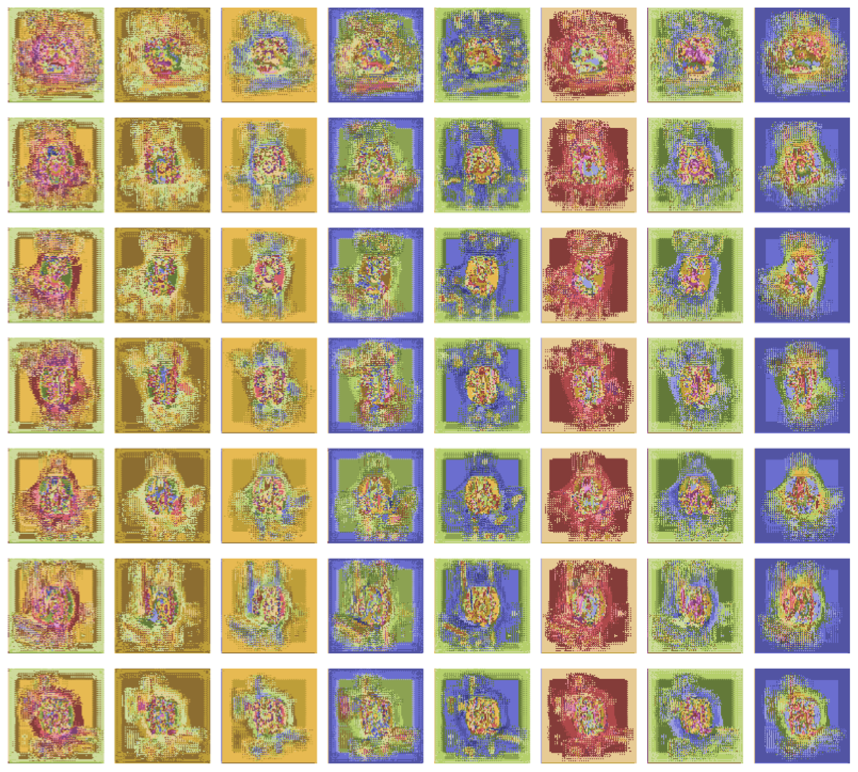}
    \caption{\textbf{Visualization of the BEV latent features.}}
    \label{fig:bev}
\end{figure}
\mypara{Visualization of the BEV latent features.}
As shown in \cref{fig:bev}, we present the visualization of the BEV latent features, demonstrating their correspondence to street layouts and their effective discretization by the vector quantization module.






\end{document}